\documentclass[10pt,twocolumn,letterpaper]{article}

\usepackage{iccv}
\usepackage{times}
\usepackage{epsfig}
\usepackage{graphicx}
\usepackage{amsmath}
\usepackage{amssymb}
\usepackage{enumerate}
\usepackage{enumitem}
\usepackage{bm}
\usepackage[linesnumbered,lined,boxed,commentsnumbered]{algorithm2e}
\usepackage{color}
\PassOptionsToPackage{hyphens}{url}
\definecolor{mypink}{RGB}{255,135,180}
\definecolor{myyellow}{RGB}{255,200,0}
\definecolor{mygreen}{RGB}{0,100,0}
\definecolor{mygold}{RGB}{255,185,0}
\definecolor{mymaroon}{RGB}{128,0,0}
\definecolor{myaquamarine}{RGB}{120,200,190}
\definecolor{myturquoise}{RGB}{64,224,208}
\definecolor{mymagenta}{RGB}{255,0,255}
\definecolor{myviolet}{RGB}{238,130,238}

\usepackage[breaklinks=true,bookmarks=false]{hyperref}
\usepackage{hyperref}

\iccvfinalcopy 

\def\iccvPaperID{1820} 

\ificcvfinal\fi

\begin{document}

\title{Weakly Supervised Energy-Based Learning for Action Segmentation}

\author{Jun Li\\
Oregon State University\\
{\tt\small liju2@oregonstate.edu}
 \and  
 Peng Lei\thanks{The work was done at the Oregon State University before Peng Lei joined Amazon. \textsuperscript{\dag}The code is available at \textcolor{magenta}{https://github.com/JunLi-Galios/CDFL}.}\\
 Amazon.com Services, Inc.\\
 {\tt\small  leipeng@amazon.com}
 \and 
 Sinisa Todorovic\\
 Oregon State University\\
 {\tt\small sinisa@oregonstate.edu}
 }
\newcommand\model{Constrained Discriminative Forward Loss}
\newcommand\abbrmodel{CDFL}

\maketitle
\ificcvfinal\thispagestyle{empty}\fi

\begin{abstract}
   This paper is about labeling video frames with action classes under weak supervision in training, where we have access to a temporal ordering of actions, but their start and end frames in training videos are unknown. Following prior work, we use an HMM grounded on a Gated Recurrent Unit (GRU) for frame labeling. Our key contribution is a new constrained discriminative forward loss (CDFL) that we use for training the HMM and GRU under weak supervision. While prior work typically estimates the loss on a single, inferred video segmentation, our CDFL discriminates between the energy of all valid and invalid frame labelings of a training video. A valid frame labeling satisfies the ground-truth temporal ordering of actions, whereas an invalid one violates the ground truth. We specify an efficient recursive algorithm for computing the CDFL in terms of the logadd function of the segmentation energy. Our evaluation on action segmentation and alignment gives superior results to those of the state of the art on the benchmark Breakfast Action, Hollywood Extended, and 50Salads datasets.\textsuperscript{\dag}
   %
\end{abstract}

\section{Introduction}\label{sec:Introduction}
This paper presents an approach to weakly supervised action segmentation by labeling video frames with action classes. Weak supervision means that in training our approach has access only to the temporal ordering of actions, but their ground-truth start and end frames are not provided. This is an important problem with a wide range of applications, since the more common fully supervised action segmentation typically requires expensive manual annotations of action occurrences in every video frame.


Our fundamental challenge is that the set of all possible segmentations of a training video may consist of multiple distinct {\em valid} segmentations that satisfy the provided ground-truth ordering of actions, along with {\em invalid} segmentations that violate the ground truth. It is not clear how to estimate loss (and subsequently train the segmenter) over multiple valid segmentations.

{\bf Motivation:}  Prior work \cite{huang2016connectionist, richard2017weaklysupervised, richard2017weakly, ding2018weakly, richard2018neuralnetwork} typically uses a temporal model (e.g., deep neural network, or HMM) to infer a {\em single}, valid, optimal video segmentation, and takes this inference result as a pseudo ground truth for estimating the incurred loss. However, 
a particular training video may exhibit a significant variation (not yet captured by the model along the course of training), which may negatively affect estimation of the pseudo ground truth, such that the inferred action segmentation is significantly different from the true one. In turn, the loss estimated on the incorrect pseudo ground truth may corrupt  training by reducing, instead of maximizing, the discriminative margin between the ground truth and other valid segmentations.
In this paper, we seek to alleviate these issues.

{\bf Contributions:} Prior work shows that a statistical language model is useful for weakly supervised learning and modeling of video sequences  \cite{lin2017ctc, koller2017re, richard2016temporal, richard2018neuralnetwork, chang2019d3tw}. Following \cite{richard2018neuralnetwork}, we also adopt a Hidden Markov Model (HMM) grounded on a Gated Recurrent Unit (GRU) \cite{chung2014empirical} for labeling video frames. The major difference is that we do not generate a unique pseudo ground truth for training. Instead, we efficiently account for all candidate segmentations of a training video when estimating the loss. To this end, we formulate a new Constrained Discriminative Forward Loss (CDFL) as a difference between the energy of valid and invalid candidate video segmentations. In comparison with prior work, the CDFL improves robustness of our training, because minimizing the CDFL amounts to maximizing the discrimination margin between candidate segmentations that satisfy and violate ground truth, whereas prior work solely optimizes a score of the inferred single valid segmentation. Robustness of training is further improved when the CDFL takes into account only hard invalid segmentations whose edge energy is lower than that of valid ones.  Along with the new CDFL formulation, our key contribution is a new recursive algorithm for efficiently estimating the CDFL in terms of the {\em logadd} function of the segmentation energy.

\begin{figure}[t]
\centering
\includegraphics[width=0.5\textwidth]{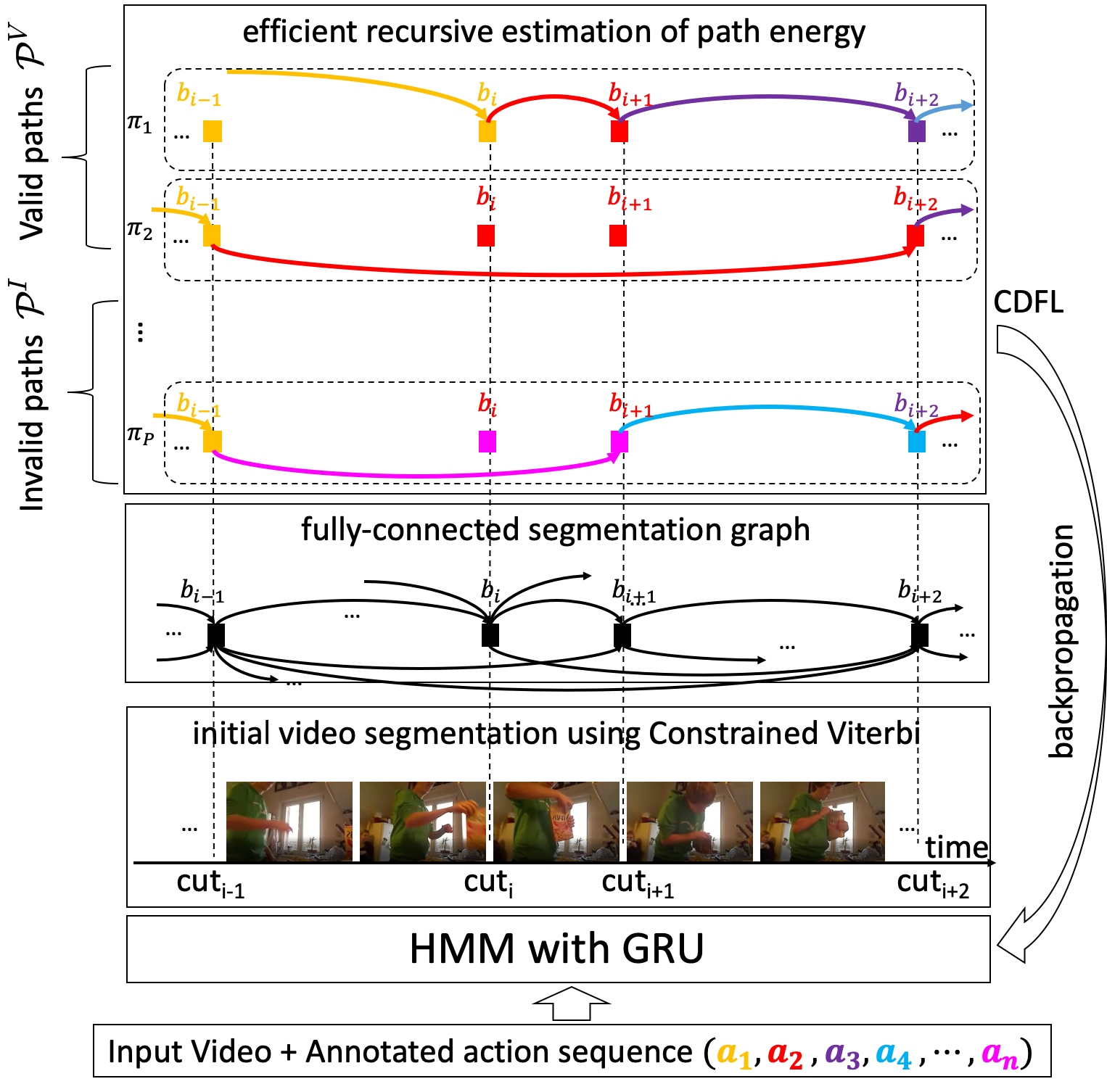}
\caption{{\bf Our weakly supervised training}: 
For a training video, we first estimate candidate segmentation cuts using a Hidden Markov Model (HMM) grounded on a Gated Recurrent Unit (GRU), and then build a fully connected segmentation graph whose paths represent candidate action segmentations (colors mark different action classes along the paths).  Then, we efficiently compute the Constrained Discriminative Forward Loss (CDFL) in terms of accumulated energy of all valid and invalid paths in the graph for our end-to-end training. (best seen in color)}
\label{fig:Introduction}
\vspace{-4mm}
\end{figure}

{\bf Our Approach:} Fig.~\ref{fig:Introduction} shows an overview of our weakly supervised training of the HMM with GRU that consists of two steps. In the first step, we run a constrained Viterbi algorithm for HMM inference on a given training video so the resulting segmentation is valid. This initial video segmentation is used for efficiently building a fully connected segmentation graph aimed at representing alternative candidate segmentations. In this graph, nodes represent segmentation cuts of the initially inferred segmentation -- i.e., video frames where one action ends and a subsequent one starts -- and edges represent video segments between every two temporally ordered cuts. For improving action boundary detection, we further augment the initial set of nodes with video frames that are in a vicinity of every cut, as well as the initial set of edges with corresponding temporal links between the added nodes. Directed paths of such a fully connected graph explicitly represent many candidate action segmentations, beyond the initial HMM's inference.

The second step of our training efficiently computes a total energy score of frame labeling along {\em all paths} in the segmentation graph. Efficiency comes from our novel recursive estimation of the segmentation energy, where we exploit the accumulative property of the {\em logadd} function. A difference of the accumulated energy of action labeling along the valid and invalid paths is used to compute the CDFL. In this paper, we also consider several other loss formulations expressed in terms of the energy of valid and invalid paths. The loss is then used for training HMM parameters and back-propagated to the GRU for end-to-end training.

For inference on a test video, as in the first step of our training, we use a constrained Viterbi algorithm to perform the HMM inference 
which will satisfy at least one action sequence seen in training. 
Then, we use this initial video segmentation as an anchor for building the segmentation graph that comprises paths with finer action boundaries. Our output is the MAP path in the graph. 

For evaluation, we consider the tasks of action segmentation and action alignment, where the latter provides additional information on the temporal ordering of actions in the test video.  For both tasks on the Breakfast Action dataset \cite{kuehne2014language}, Hollywood Extended dataset \cite{bojanowski2014weakly}, and 50-Salads dataset \cite{stein2013combining}, we outperform the state of the art.


In the following, Sec.~\ref{sec:Related Work} reviews  related work, Sec.~\ref{sec:HMM} formulates our HMM and Constrained Viterbi for action segmentation, Sec.~\ref{sec:SegmentationGraph} describes how we construct the segmentation graph, Sec.~\ref{sec:CDFL} specifies our CDFL and related loss functions, and Sec.~\ref{sec:Experiment} presents our evaluation.

\section{Related Work}\label{sec:Related Work}
This section reviews closely related work on weakly supervised action segmentation and Graph Transformer Networks. While a review of fully supervised action segmentation \cite{yeung2016end, lea2016segmental, rene2017temporal, lei2018temporal} is beyond our scope, it is worth mentioning that our approach uses the same recurrent deep models for frame labeling as in \cite{singh2016multi,yeung2016end, ding2017tricornet}. Also, our approach is motivated by \cite{kuehne2016end, richard2016temporal} which integrate HMMs  and modeling of action length priors  within a deep learning architecture.



{\bf Weakly supervised action segmentation} has recently made much progress \cite{stein2013combining, kuehne2014language, richard2017weakly, ding2018weakly, richard2018neuralnetwork}.  For example, Extended Connectionist Temporal Classification (ECTC) addresses action alignment under the constraint of being consistent with frame-to-frame visual similarity \cite{huang2016connectionist}. Also,  action segmentation has been addressed with a convex relaxation of discriminative clustering, and efficiently solved with the conditional gradient (Frank-Wolfe) algorithm  \cite{bojanowski2014weakly}. Other approaches use a local action model and a global temporal alignment model that are alternatively trained  \cite{richard2017weaklysupervised, richard2017weakly}. Some methods initially predict a video segmentation with a temporal convolutional network, and then iteratively refine the action boundaries \cite{ding2018weakly}. Other approaches first  generate pseudo-ground-truth labels for all video frames, e.g., with the Viterbi algorithm \cite{richard2018neuralnetwork}, and then train a classifier on these frame labels by minimizing the standard cross entropy loss. Finally, \cite{richard2018actionsets} addresses a different weakly supervised setting from ours when the ground truth provides only a set of actions present without their temporal ordering .

All these approaches base their learning and prediction on estimating a penalty or probability of labeling individual frames. In contrast, we use an energy-based framework with the following differences. First, in training, we minimize the total energy of valid paths in the segmentation graph rather than optimize labeling probabilities of each frame. Second, instead of considering a single optimal valid path in the segmentation graph, we specify a loss function in terms of {\em all} valid paths. Hence, the Viterbi-initialized training on pseudo-labels of frames \cite{richard2018neuralnetwork} represents a special case of our training done only for one valid path. In addition, our loss enforces discriminative training by accounting for invalid paths in the segmentation graph. Unlike \cite{chang2019d3tw} that randomly selects invalid paths, we efficiently account for all hard invalid paths in training.
Finally, our training is not iterative as in \cite{richard2017weaklysupervised, richard2017weakly}, and does not require iterative refinement of action boundaries as in \cite{ding2018weakly}. 

Our \abbrmodel\ extends the loss used for training of the Graph Transformer Network (GTN) \cite{lecun1998gradient, le1997reading, bottou2005graph, collobert2011deep}. 
To the best of our knowledge, the GTN has been used only for text parsing, and never for action segmentation. In comparison with the GTN training, we significantly reduce complexity by building the video's segmentation graph. Also, while the loss used for training the GTN accounts for both valid and invalid text parses, it cannot handle the special case when valid parses have lower scores than invalid ones. In contrast, our CDFL effectively accounts for the energy of valid and invalid paths, even when valid paths have significantly lower energy than invalid paths in the segmentation graph.

\section{Our Model for Action Segmentation}\label{sec:HMM}
\noindent
{\bf Problem Setup:} For each training video of length $T$, we are given unsupervised frame-level features, $\bm{x}_{1:T} = [x_1,x_2,...,x_T]$, and the ground-truth ordering of action classes $\bm{a}_{1:N} = [a_1, a_2, ..., a_N]$, also referred to as the transcript.  $N$ is the length of the annotation sequence, and $a_n$ is $n$th action class in $\bm{a}_{1:N}$  that belongs to the set of $K$ action classes, $a_n\in\mathcal{A}=\{1,2,...,K\}$. Note that $T$ and $N$ may vary across the training set, and that there may be more than one occurrences of the same action class spread out in $\bm{a}_{1:N}$ (but of course $a_n\ne a_{n+1}$).


In inference, given frame features $\bm{x}_{1:T}$ of a video,  our goal is to find an optimal segmentation $(\hat{\bm{a}}_{1:\hat{N}}, \hat{\bm{l}}_{1:\hat{N}})$,  where $\hat{N}$ is the predicted length of the action sequence, and  $\hat{\bm{l}}_{1:\hat{N}} = [\hat{l}_1,\hat{l}_2,\cdots, \hat{l}_{\hat{N}}]$ includes the predicted number of video frames $\hat{l}_n$ occupied by the predicted action $\hat{a}_n$.

{\bf The Model:} We use an HMM to model the posterior distribution of a video segmentation $(\bm{a}_{1:N}, \bm{l}_{1:N})$ given  $\bm{x}_{1:T}$ as
\begin{equation}
\begin{array}{l}
p(\bm{a}_{1:N}, \bm{l}_{1:N}|\bm{x}_{1:T})   \\ \quad \quad \propto p(\bm{x}_{1:T}|\bm{a}_{1:N}, \bm{l}_{1:N})p(\bm{l}_{1:N}|\bm{a}_{1:N})p(\bm{a}_{1:N}) , \\ \quad \quad
   = \displaystyle \left(\prod_{t=1}^{T} p(x_t|a_{n(t)})\right) \left(\prod_{n=1}^{N}  p(l_n| a_n) \right) p(\bm{a}_{1:N}).
   \end{array}
   \label{eq:Markov}
\end{equation}
In \eqref{eq:Markov}, the likelihood $p(x_t|a)$ is estimated as
\begin{equation}
    p(x_t|a) \propto \frac{p(a|x_t)}{p(a)},
    \label{eq:framelikelihood}
\end{equation}
where $p(a|x_t)$ is the GRU's softmax score for action $a\in \mathcal{A}$ at frame $t$, and the prior distribution of action classes $p(a)$ is an normalized frame frequency of action occurrences in the training dataset. The likelihood of action length is modeled as a class-dependent Poisson distribution 
\begin{equation}
    p(l|a) = \frac{\lambda_{a}^{l} }{l!}e^{-\lambda_{a}} , 
    \label{eq:Poisson}
\end{equation}
where $\lambda_{a}$ is the mean length for class $a\in \mathcal{A}$. Finally, the joint prior $p(\bm{a}_{1:N})$ is a constant if the transcript $\bm{a}_{1:N}$ exists in the training set; otherwise, $p(\bm{a}_{1:N})=0$. The same modeling formulation was well-motivated and used in state of the art \cite{richard2018neuralnetwork}.

{\bf Constrained Viterbi Algorithm:} Given a training video, we first find an optimal valid action segmentation $(\hat{\bm{a}}_{1:\hat{N}}, \hat{\bm{l}}_{1:\hat{N}})$ by maximizing \eqref{eq:Markov} with a constrained Viterbi algorithm, which ensures that $\hat{\bm{a}}_{1:\hat{N}}$ is equal to the annotated transcript, $\hat{\bm{a}}_{1:\hat{N}}=\bm{a}_{1:N}$. Similarly, for inference on a test video, we first perform the constrained Viterbi algorithm against all transcripts $\{\bm{a}_{1:N}\}$ seen in training, i.e., ensure that the predicted $\hat{\bm{a}}_{1:\hat{N}}$ has been seen at least once in training. Thus, the initial step of our inference on a training or test video is the same as in \cite{richard2018neuralnetwork}.

Our key difference from \cite{richard2018neuralnetwork}, is that we use the initial $(\hat{\bm{a}}_{1:\hat{N}}, \hat{\bm{l}}_{1:\hat{N}})$ to efficiently build a fully connected segmentation graph of the video, as explained in Sec.~\ref{sec:SegmentationGraph}. 
Importantly, in training, the segmentation graph is not constructed to find a more optimal video segmentation that improves upon the initial prediction. Instead, the graph is used to efficiently account for all valid and invalid segmentations. 

Given a video $\bm{x}_{1:T}$ and a transcript $\bm{a}_{1:N}$, the constrained Viterbi algorithm recursively maximizes the posterior in \eqref{eq:Markov} such that the first $n$ action labels of the transcript $\bm{a}_{1:n}=[a_1,...,a_n]\subseteq \bm{a}_{1:N}$ are respected at time $t$:
\begin{equation}
\begin{array}{l}
\displaystyle
  p(\bm{a}_{1:n},\hat{\bm{l}}_{1:n}|\bm{x}_{1:t}) = \left.\max_{t',\; t'<t}\right\{ p(\bm{a}_{1:n-1},\hat{\bm{l}}_{1:n-1}|\bm{x}_{1:t'})\\ \displaystyle \quad \quad   \cdot \left.\left(\prod_{s = t'}^{t} p(x_s|a_{n(s)})\right) \cdot p(l_n|a_n) \cdot p(\bm{a}_{1:n})\right\},
  \end{array}
\label{eq:Viterbi}
\end{equation}
where $l_n = t-t'$. We set $p(\cdot|\bm{x}_{1:0}) = 1$, and $p(\bm{a}_{1:n}) = \kappa$, where $\kappa>0$ is a constant.

\section{Constructing the Segmentation Graph}\label{sec:SegmentationGraph}

Given a video $\bm{x}_{1:T}$, we first run the constrained Viterbi algorithm to obtain an initial video segmentation  $(\hat{\bm{a}}_{1:\hat{N}}, \hat{\bm{l}}_{1:\hat{N}})$. For simplicity, in the following, we ignore the symbol $\hat{~}$. This initial segmentation is characterized by $N+1$ cuts, $\bm{b}_{1:N+1}=[b_1,\dots, b_{N+1}]$, i.e., video frames where previous action ends and the next one starts including the very first frame $b_1$ and last frame $b_{N+1}$ at time $T$. 

We use these cuts to anchor our construction of the fully connected segmentation graph, $G=(\mathcal{V},\mathcal{E},\mathcal{W})$, where $\mathcal{V}=\{\bm{b}_{1:N+1}\}$ is the set of nodes, $\mathcal{E}$ is the set of directed edges linking every two temporally ordered nodes, and $\mathcal{W}$ are the corresponding edge weights. 

\begin{figure}
\centering
\includegraphics[width=0.49\textwidth]{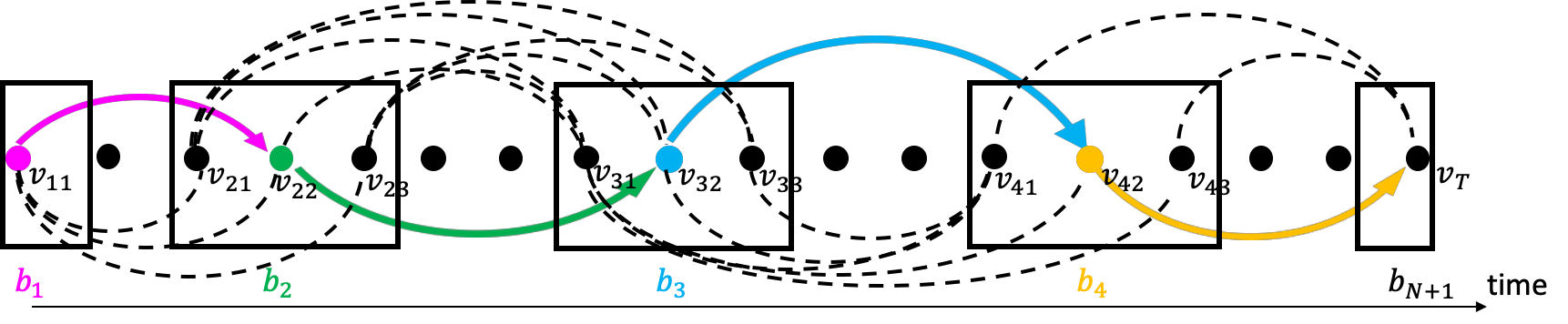}
\caption{{\bf Building the segmentation graph} $G$ (best seen in color). The initial nodes of $G$ represent segmentation cuts $b_{n}$ obtained in the Constrained Viterbi (the predicted action classes are marked with different colors). Each $b_{n}$ generates additional vertices $b_n=\{v_{ns}\}$ representing neighboring video frames within a window  centered at $b_n$ (the black rectangles), and corresponding new edges $(v_{ns},v_{n's'})$ (the dashed lines) between all temporally ordered pairs of vertices in $G$. For clarity, we show only a few edges. $G$ has exponential many  paths, each representing a candidate action segmentation.}
\label{fig:NodeAugment}
\end{figure}

Some of the estimated cuts in $\bm{b}_{1:N+1}$ may be false positives or may not exactly coincide with the true cuts. To improve action boundary detection, we augment the initial $\mathcal{V}$ with nodes representing neighboring video frames of each cut $b_{n}$ within a temporal window of length $\Delta$ centered at $b_{n}$, as illustrated in Fig.~\ref{fig:NodeAugment}. For the first and last frames, we set $\Delta=1$. Thus, each $b_{n}$ can be viewed as a hyper-node comprising additional vertices in $G$, $\mathcal{V}=\{b_n=\{v_{n1},\cdots,v_{ni},\cdots,v_{n\Delta}\}:n=1,\dots,N+1\}$, and accordingly additional  edges $\mathcal{E}=\{(v_{ni},v_{n'i'}): n\le n', \; i<i'\}$.  In the following, we simplify notation for vertices $v_{ni}\to v_i \in \mathcal{V}$, and edges  $(v_{ni},v_{n'i'}) \to e_{ii'}=(v_i,v_{i'})$.

Each edge $e_{ii'}$ is assigned a weight vector  $\bm{w}_{ii'}=[w_{ii'}(a)]$, where  $w_{ii'}(a)$ is defined as the energy of labeling the video segment  $(v_i,v_{i'})$ with action class $a\in\mathcal{A}$:
\begin{equation}
w_{ii'}(a) = \sum_{t\in(v_i,v_{i'})} -\log p(a|x_t),
\label{eq:edgeweight}
\end{equation}
where $p(a|x_t)$ is the GRU's softmax score for action $a$ at frame $t$.

$G$ comprises exponentially many directed paths $\mathcal{P}=\{\pi\}$, where each $\pi$ represents a particular video segmentation. In each $\pi$, every edge $e_{ii'}$ gets assigned only one action class $a_{ii'}^{\pi}\in\mathcal{A}$. Thus, the very same edge with $K$ different class assignments belongs to $K$ distinct paths in $\mathcal{P}$. We compute the energy of a path as%
\begin{equation}
    E_{\pi}= \sum_{e_{ii'}\in\pi}  w_{ii'}(a_{ii'}^{\pi}).
\label{eq:energy}
\end{equation}

A subset of valid paths $\mathcal{P}^V\subset \mathcal{P}$ satisfies the given transcript. The other paths are invalid, $\mathcal{P}^I= \mathcal{P}\setminus \mathcal{P}^V$.

In the next section, we explain how to efficiently compute a total energy score of the exponentially many paths in $\mathcal{P}$ for estimating our loss in training.

\section{\model}\label{sec:CDFL}

In this paper, we study {\em three} distinct loss functions, defined in terms of a total energy score of paths in $G$. As there are exponentially many paths in $G$, our key contribution is the algorithm for efficiently estimating their total energy.  Below, we specify our three loss functions ordered by their complexity. As we will show in Sec.~\ref{sec:Experiment}, we obtain the best performance when using the CDFL in training.

\subsection{Forward Loss}

We define a forward loss,  $L_\text{F}$, in terms of a total energy of all valid paths using the standard {\em logadd} function as
\begin{equation}\label{eq:ForwardTraining}
L_\text{F} = -\log(\sum_{\pi\in\mathcal{P}^V}\exp(-E_{\pi})),
\end{equation}
where energy of a path $E_{\pi}$ is given by \eqref{eq:energy}. As there are exponentially many paths in $\mathcal{P}^V$, we cannot directly compute $L_\text{F}$ as specified in (\ref{eq:ForwardTraining}).  Therefore, we derive a novel recursive algorithm for accumulating the energy scores of edges along multiple paths, as specified below.

We begin by defining the {\em logadd} function as
\begin{equation}
    \text{logadd}(a,b) = -\log(\exp(-a) + \exp(-b)).
\end{equation}
Note that the {\em logadd} function is commutative and associative, so  it can be defined on  a set $S$ in a recursive manner:
\begin{equation}
    \text{logadd}(S) = \text{logadd}(S\backslash  \{x\}, x),
\end{equation}
where $x$ is an element in $S$.
Therefore, the forward loss given by \eqref{eq:ForwardTraining} can be expressed as
\begin{equation}
L_\text{F} = \text{logadd}(\{E_{\pi}:\pi \in \mathcal{P}^V\}).
\end{equation}
Below, we simplify notation as $L_\text{F} = \text{logadd}(\mathcal{P}^V)$.

\begin{algorithm}[t]
\SetAlgoLined
\KwIn{$G,\; \bm{b}_{1:N+1},\; \bm{a}_{1:N}$}
\KwOut{Forward loss $ L_\text{F} =\ell_T(\bm{a})$}
Initialization: $\ell_0(\cdot) = 0$ \;
\For{$n= 1$ \KwTo $N$}{
\For{$i'$ {\normalfont in the neighborhood of}  $b_n$}{
$\ell_{i'}(\bm{a}_{1:n}) = \infty$\;
\For{$i$  {\normalfont in the neighborhood of} $b_{n-1}$}{
$\text{temp} = \ell_{i}(\bm{a}_{1:n-1}) + w_{ii'}(a_n)$\;
$\ell_{i'}(\bm{a}_{1:n}) = \text{logadd}(\ell_{i'}(\bm{a}_{1:n}), \text{temp})$\;
}
}
}
\caption{Computing the Forward loss  $L_\text{F}$. }
\label{alg:ForwardLoss}
\end{algorithm}

We recursively compute the energy score $\ell_{i'}(\bm{a}_{1:n})$ of a path that ends at node $i'$ and covers first $n$ labels of the ground truth   $\bm{a}_{1:n}=[a_1,...,a_n]\subseteq \bm{a}_{1:N}$ in terms of the {\em logadd} scores $\ell_{i}(\bm{a}_{1:n-1})$ of all valid paths that end at node $i$, $i<i'$, and cover first $n-1$ labels as
\begin{equation}\label{eq:ELL}
\ell_{i'}(\bm{a}_{1:n}) = \text{logadd}(\{ \ell_{i}(\bm{a}_{1:n-1}) + w_{ii'}(a_n): i<i'\}).
\end{equation}
To prove \eqref{eq:ELL}, suppose that
\begin{align}
\ell_{i}(\bm{a}_{1:n-1}) & = \text{logadd}(\{E_{\pi_{i}}:\pi_{i} \in \mathcal{P}^V\}) \nonumber \\
 &= -\log(\sum_{\pi_i\in\mathcal{P}^V}\exp(-E_{\pi_i})), 
\end{align}
where $\pi_i$ is a path that ends at $i$ with a transcript of $\bm{a}_{1:n-1}$. Then, we have
\begin{equation}
\begin{array}{lcl}
\ell_{i'}(\bm{a}_{1:n}) &=& \text{logadd}(\{ \ell_{i}(\bm{a}_{1:n-1}) + w_{ii'}(a_n): i<i'\}).\\
&=&\displaystyle -\log(\sum_{i<i^{'}}\sum_{\pi_i^{'}\in\mathcal{P}^V}\exp(-E_{\pi_i}-w_{ii'}(a_n))) \\
&=&-\log(\sum_{\pi_i^{'}\in\mathcal{P}^V}\exp(-E_{\pi_i^{'}})) \\
&=&\text{logadd}(\{E_{\pi_{i^{'}}}:\pi_{i^{'}} \in \mathcal{P}^V\}).
\end{array}
\end{equation}
where $\pi_{i'}$ is a path that ends at $i'$ with a transcript of $\bm{a}_{1:n}$.
For a training video with length $T$ and ground-truth constraint sequence $\bm{a}_{1:N}$, we define
\begin{equation}\label{eq:Viterbi_F}
  L_\text{F} = \ell_T(\bm{a}_{1:N}).
\end{equation}
The recursive algorithm for computing $L_\text{F}$ is presented in Alg.~\ref{alg:ForwardLoss}. It is worth noting that in a special case of Alg.~\ref{alg:ForwardLoss}, when we take only the initial segmentation cuts $\bm{b}_{1:N+1}$ as nodes of $G$ (i.e., the window size $\Delta=0$), the forward loss is equal to the training loss used in \cite{richard2018neuralnetwork}.

\begin{algorithm}[t]
\SetAlgoLined
\KwIn{$G,\; \bm{b}_{1:N+1}$}
\KwOut{$\text{logadd}(\mathcal{P})=\ell_{T}$}
Initialization: $\ell_0 = 0$ \;
\For{$n= 1$ \KwTo $N$}{
\For{$i'$ {\normalfont in the neighborhood of}  $b_n$}{
$\ell_{i'} = \infty$\;
\For{$i$  {\normalfont in the neighborhood of} $b_{n-1}$}{
\For{$a \in \mathcal{A}$}{
$\ell_{i'} = \text{logadd}(\ell_{i'}, \ell_{i} + w_{ii'}(a))$\;
}
}
}
}
\caption{Computing the logadd score of all paths in $\mathcal{P}$, for  the discriminative forward loss $L_\text{DF}$. }
\label{alg:DF}
\end{algorithm}

\subsection{Discriminative Forward Loss}
We also consider the Discriminative Forward Loss, $L_\text{DF}$, which extends $L_\text{F}$ by additionally accounting for invalid paths in $G$:
\begin{equation}
  L_\text{DF} = \text{logadd}(\mathcal{P}^V) - \alpha\; \text{logadd}(\mathcal{P}),
  \label{eq:DF}
\end{equation}
where $\text{logadd}(\mathcal{P})$ aggregates  a total energy of all paths in $G$, and  $\alpha>0$ is a regularization factor that controls the relative importance of the valid and invalid paths for $L_\text{DF}$. Alg.~\ref{alg:DF} summarizes our recursive algorithm for computing $\text{logadd}(\mathcal{P})$ in \eqref{eq:DF}, whereas Alg.~\ref{alg:ForwardLoss} shows how to compute $\text{logadd}(\mathcal{P}^V)$ in \eqref{eq:DF}.  

One advantage of $L_\text{DF}$ over $L_\text{F}$ is that minimizing $L_\text{DF}$ amounts to  maximizing the decision margin between the valid and invalid paths. However, a potential shortcoming of $L_\text{DF}$ is that valid paths might have little effect in \eqref{eq:DF}. In the case, when the energy of valid paths dominates the total energy of all paths, the former gets effectively subtracted in \eqref{eq:DF}, and hence has very little effect on learning. 

Moreover, we observe that in some cases the back-propagation of $L_\text{DF}$ is dominated by the invalid paths. This can be clearly seen from the following derivation. We compute the gradient $\nabla  L_\text{DF}$  as
\begin{equation}
\begin{array}{lcl}
\nabla  L_\text{DF} &=& \nabla \text{logadd}(\mathcal{P}^V) -\alpha\; \nabla\text{logadd}(\mathcal{P}),  \\

                    &=&  c_1\sum_{\pi\in\mathcal{P}^V}\exp(-E_{\pi})\nabla E_{\pi} \\
                    &&- c_2\sum_{\pi\in\mathcal{P}^I}\exp(-E_{\pi})\nabla E_{\pi},\\
\end{array}
\label{eq:gradientDF}
\end{equation}
where
\begin{equation}
\begin{array}{lcl}
c_1 &=& \frac{(1-\alpha)\sum_{\pi\in\mathcal{P}^V} \exp(-E_{\pi})+\sum_{\pi\in\mathcal{P}^I} \exp(-E_{\pi})}{(\sum_{\pi\in\mathcal{P}^V} \exp(-E_{\pi}))(\sum_{\pi\in\mathcal{P}} \exp(-E_{\pi})) },\\
                    c_2&=&\frac{\alpha}{\sum_{\pi\in\mathcal{P}} \exp(-E_{\pi})}.
\end{array}
\label{eq:c1c2}
\end{equation}
From \eqref{eq:gradientDF}--\eqref{eq:c1c2}, we note that in the case of $\alpha \to 1$, the backpropagation will be dominated by the invalid paths, whereas there would be no effect for invalid paths in training if $\alpha = 0$. Sec.~\ref{sec:Experiment} presents how different choices of $\alpha$ affect our performance.

In the next section, we define the constrained discriminative forward loss to address this issue.

\begin{algorithm}[t]
\SetAlgoLined
\KwIn{$G,\; \bm{b}_{1:N+1},\; \bm{a}_{1:N}$}
\KwOut{ $\text{logadd}(\mathcal{P}^{I^c})=\ell_T$}
Initialization: $\ell_0 = 0$ \;
\For{$n= 1$ \KwTo $N$}{
\For{$i'$ {\normalfont in the neighborhood of}  $b_n$}{
$\ell_{i'} = \infty$\;
\For{$i$  {\normalfont in the neighborhood of} $b_{n-1}$}{
\For{$a \in \mathcal{A}$}{
$\text{temp}=\ell_{i}$\;
\If{$w_{ii'}(a) < w_{ii'}(a_n)$}{
$\text{temp}=\ell_{i}+ w_{ii'}(a)$
}
$\ell_{i'} = \text{logadd}(\ell_{i'}, \text{temp})$\;
}
}
}
}
\caption{Computing the logadd score of a subset of invalid paths $\mathcal{P}^{I_c}$, for estimating the constrained discriminative forward loss $L_\text{CDF}$.}
\label{alg:CDF}
\end{algorithm}


\subsection{Constrained Discriminative Forward Loss}

We define the CDFL as
\begin{equation}
  L_\text{CDF} = \text{logadd}(\mathcal{P}^V) - \text{logadd}(\mathcal{P}^{I_c}),
  \label{eq:CDFLoss}
\end{equation}
where $\mathcal{P}^{I_c}$ consists of a subset of invalid paths in $G$, where each edge $e_{ii'}$ gets assigned an action class $a$ such that its weight $w_{ii'}(a) < w_{ii'}(a_n)$, where $a_n\ne a$ is the pseudo ground truth class for $e_{ii'}$. This constraint effectively addresses the aforementioned issue when the valid paths have significantly lower energy than the invalid paths. 
Alg.~\ref{alg:CDF} summarizes our recursive algorithm for computing $\text{logadd}(\mathcal{P}^{I_c})$ in \eqref{eq:CDFLoss}, whereas Alg.~\ref{alg:ForwardLoss} shows how to compute $\text{logadd}(\mathcal{P}^V)$ in \eqref{eq:CDFLoss}. 

As $L_\text{DF}$ accounts for the invalid paths,  $L_\text{CDF}$ further accounts for the hard invalid paths. Therefore, the model robustness is further improved by minimizing $L_\text{CDF}$ which amounts to maximizing the decision margin between the valid and hard invalid paths.

\begin{table}[b]
\begin{center}
\begin{tabular}{|c|c| c| c| c|}
\hline {\bf Breakfast}  &Mof & Mof-bg & IoU & IoD\\
\hline OCDC\cite{bojanowski2014weakly}& 8.9 & - & - & -\\
\hline CTC\cite{huang2016connectionist}& 21.8 & - & - & -\\
\hline HTK \cite{kuehne2016end}& 25.9 & - & 9.8 & -\\
\hline ECTC \cite{huang2016connectionist}& 27.7 & - & - & -\\
\hline HMM/RNN \cite{richard2017weakly}& 33.3 & - & - & -\\
\hline TCFPN \cite{ding2018weakly} & 38.4 & 38.4 & 24.2 & 40.6\\
\hline NN-Viterbi \cite{richard2018neuralnetwork} & 43.0 & - & - & -\\
\hline D3TW \cite{chang2019d3tw} & 45.7 & - & - & -\\
\hline Our \abbrmodel & {\bf 50.2} & {\bf 48.0} & {\bf 33.7} & {\bf 45.4} \\
\hline
\hline {\bf Hollywood Ext}  &Mof & Mof-bg & IoU & IoD\\
\hline HTK \cite{kuehne2016end}& 33.0 & - & 8.6 & -\\
\hline HMM/RNN \cite{richard2017weakly}& - & - & 11.9 & -\\
\hline TCFPN \cite{ding2018weakly} & 28.7 & 34.5 & 12.6 & 18.3\\
\hline D3TW \cite{chang2019d3tw} & 33.6 & - & - & -\\
\hline Our \abbrmodel & {\bf 45.0} & {\bf 40.6} & {\bf 19.5} & {\bf 25.8}\\
\hline
\hline {\bf 50Salads}  &Mof & Mof-bg & IoU & IoD\\
\hline CTC\cite{huang2016connectionist}& 11.9 & - & - & -\\
\hline HTK \cite{kuehne2016end}& 24.7 & - & - & -\\
\hline HMM/RNN \cite{richard2017weakly}& 45.5 & - & - & -\\
\hline NN-Viterbi \cite{richard2018neuralnetwork} & 49.4 & - & - & -\\
\hline Our \abbrmodel & {\bf 54.7} & {\bf 49.8} & {\bf 31.5} & {\bf 40.4} \\
\hline
\end{tabular}
\end{center}
\caption{Action segmentation evaluations on Breakfast, Hollywood Ext and 50Salads. The dash means no results reported by prior work.}
\label{table:result}
\end{table}
\setlength{\tabcolsep}{1.4pt}

\subsection{Our Computational Efficiency}\label{sec:Effectiveness}

As summarized in Alg.~\ref{alg:ForwardLoss}--\ref{alg:CDF}, our training first runs the constrained Viterbi algorithm (see Sec.~\ref{sec:HMM}) to get the initial segmentation cuts with complexity  $O(T^2N)$ for a video of length $T$ and the ground-truth action sequence of length $N$. Then, \abbrmodel\ efficiently accumulates the energy of both valid and invalid paths in $G$ with complexity $O(\Delta^2KN)$ for the neighborhood window size $\Delta$ and the class set size $K$. Therefore, our total complexity of training is $O(T^2N + \Delta^2KN)$. 

Note that prior work \cite{richard2018neuralnetwork} also runs the Constrained Viterbi with complexity $O(T^2N)$, so relative to theirs our complexity is increased by $O(\Delta^2 KN)$. This additional complexity is significantly smaller than $O(T^2N)$ as $\Delta^2 K \ll T^2$. In our experimental evaluation, we get the best results for  $\Delta \le 20$ frames, whereas video length $T$ can go to several minutes.



\section{Results}\label{sec:Experiment}

Both action segmentation and alignment are evaluated on the Breakfast Actions \cite{kuehne2014language}, Hollywood Extended \cite{bojanowski2014weakly}, and 50Salads \cite{stein2013combining} datasets. We perform the same cross-validation strategy as the state of the art, and report our average results. 
We call our approach  \abbrmodel, trained with loss given by \eqref{eq:CDFLoss}.

{\bf Datasets.} For all datasets, we use as input the pre-processed, public, unsupervised frame-level features. The same frame features are used by \cite{huang2016connectionist, richard2017weaklysupervised, richard2017weakly, richard2018neuralnetwork}. The features are dense trajectories represented by PCA-projected Fisher vectors \cite{kuehne2016end}. {\em Breakfast} \cite{kuehne2014language} consists of $1,712$ videos of people making breakfast with $10$ cooking activities. The cooking activities are comprised of $48$ action classes. On average, every video has $6.9$ action instances, and the video length ranges from a few seconds to several minutes. {\em Hollywood Extended} \cite{bojanowski2014weakly} contains $937$ video clips from different Hollywood movies, showing $16$ action classes. Each clip contains $2.5$ actions on average. {\em 50Salads} \cite{stein2013combining} has $50$ very long videos showing $17$ classes of human manipulative gestures. On average, each video has $20$ action instances. There are $600,000$ annotated frames.

{\bf  Evaluation  Metrics.} We use the following four standard metrics, as in \cite{bojanowski2014weakly, ding2018weakly}. The mean-over-frames ({\bf Mof}) is the average percentage of correctly labeled frames. To overcome the potential drawback that frames are dominated by background class, we compute mean-over-frames without background({\bf Mof-bg}) as the average percentage of correctly labeled video frames with background frames removed. The intersection over union ({\bf IoU}) and the intersection over detection ({\bf IoD}) are computed as IoU = $|GT\cap D|/|GT\cup D|$, and IoD = $|GT\cap D|/|D|$ , where $|GT|$ denotes the extent of the ground truth segment and $|D|$ is the extent of a correctly detected action segment.

{\bf Training.} We train a single-layer GRU with 
$64$ hidden units
in $10^5$ iterations, where for each iteration one training video is randomly selected. The initial learning rate of 0.01 is decreased to 0.001 at the $60,000$th iteration. The mean action lengths  $\lambda_{a}$ in \eqref{eq:Poisson},  and the action priors $p(a)$ in \eqref{eq:framelikelihood} are estimated from the history of pseudo ground truths. Unlike \cite{richard2018neuralnetwork}, we do not use the history of pseudo ground truths for computing loss in the current iteration. Consequently, our training time per iteration is less than that of \cite{richard2018neuralnetwork}. 



\subsection{Action Segmentation}
\begin{figure}
\centering
\includegraphics[width=1.0\linewidth]{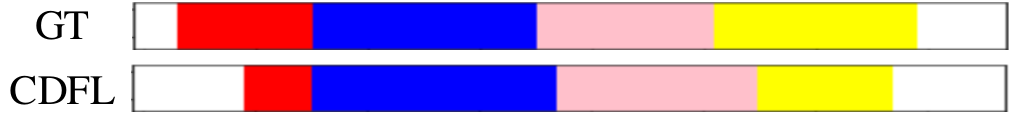}
\caption{Ground truth action sequence (\textcolor{red}{take\_cup}, \textcolor{blue}{spoon\_powder}, \textcolor{mypink}{pour\_milk}, \textcolor{myyellow}{stir\_milk}) (top) and our CDFL's action segmentation (bottom) on the sample test video $\textit{P03\_stereo01\_P03\_milk}$ from Breakfast dataset. The background frames are marked in white. \abbrmodel\ may miss the true start and end of some actions, but successfully detects the actions.}
\label{fig:prediction}
\end{figure}

 Tab.~\ref{table:result} compares \abbrmodel\ with the state of the art. From the table,  \abbrmodel\ achieves the best performance in terms of all the four metrics.
Fig.~\ref{fig:prediction} qualitatively compares the ground truth and \abbrmodel's\ output on an example test video in Breakfast dataset.
As can be seen, \abbrmodel\ typically misses the true start or end of actions by only a few frames. In general, \abbrmodel\ successfully detects most action occurrences.

\begin{table}
\begin{center}
\begin{tabular}{|c|c|c|c|c|c|c|}
\cline{1-7}
Window & \multicolumn{2}{c|}{$L_\text{F}$}&\multicolumn{2}{c|}{$L_\text{DF}$}&\multicolumn{2}{c|}{$L_\text{CDF}$}\\
\cline{2-7}
Size & Mof & IoD & Mof & IoD & Mof & IoD \\
\hline 
30 & 43.5 & 39.4 & 46.6 & 40.5 & 49.4 & 44.1 \\
\hline 
20 & 44.3 & 40.9 & 47.0 & 41.8 & {\bf 50.2} & {\bf 45.4} \\
\hline 
10 & 43.8 & 40.0 & 46.2 & 41.3 & 49.6 & 44.6\\
\hline 
0 & 43.0 & 38.7 & 45.0 & 40.2 & 48.5 & 43.5\\
\hline
\end{tabular}
\end{center}
\caption{Mof and IoD evaluations on Breakfast for different neighborhood window sizes and different losses. CDFL with neighbor-window size of 20 shows the best result.}
\label{table:Different Training Strategies}
\end{table}

{\bf  Ablation Study for Action Segmentation.} Tab.~\ref{table:Different Training Strategies} compares our action segmentation performance on Breakfast when using different sizes of the neighborhood window placed around the initial segmentation cuts (as explained in Sec.~\ref{sec:SegmentationGraph}) and different loss functions (as specified in Sec.~\ref{sec:CDFL}). From the table, training by accounting for invalid paths in $L_{\text{DF}}$ and $L_{\text{CDF}}$ gives better performance than only accounting for valid paths in $L_{\text{F}}$. In addition, considering neighboring frames for action boundary refinement within a window around the initial segmentation cuts gives better performance than taking into account only a single optimal path in the segmentation graph when the window size is $0$. The best test performance is achieved using $L_{\text{CDF}}$ with window size of $20$ in training.

\begin{figure}
\centering
\includegraphics[width=1.0\linewidth]{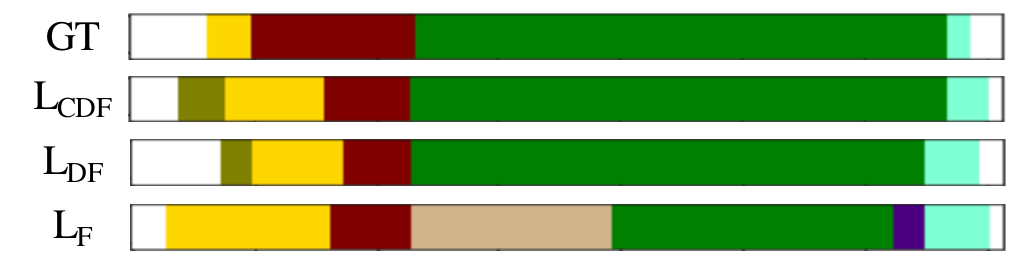}
\caption{Top-down, the rows
correspond to ground truth sequence of actions (\textcolor{mygold}{pour\_oil}, \textcolor{mymaroon}{crack\_egg}, \textcolor{mygreen}{fry\_egg}, \textcolor{myaquamarine}{put\_egg2plate}) and our action segmentations with neighbor-window size of $20$ on the sample video \textit{P03\_cam01\_P03\_friedegg} from Breakfast dataset using $L_{\text{CDF}}$, $L_{\text{DF}}$ and $L_{\text{F}}$, respectively. The background frames are marked in white. The result for $L_{\text{CDF}}$ is the best.}
\label{fig:diff_strategies}
\end{figure}

\begin{table}[t]
\begin{center}
\begin{tabular}{|c|c|c|c|c|}
\hline  window size & $\alpha=0$ & $\alpha=0.1$ & $\alpha=0.2$ & $\alpha=0.3$\\
\hline 30 & 43.5& 46.6 & 38.8 & 34.0\\
\hline 20 & 44.3& 47.0 & 40.7 & 35.5\\
\hline 10 & 43.8& 46.2 & 41.0 & 35.4\\
\hline 0 & 43.0& 45.0 & 39.1 & 33.5\\
\hline
\end{tabular}
\end{center}
\caption{Mof evaluations on Breakfast using $L_{\text{DF}}$ in training with different regularization factors and neighbor-window sizes.}
\label{table:Different regularization}
\end{table}

\begin{figure}
\centering
\includegraphics[width=1.0\linewidth]{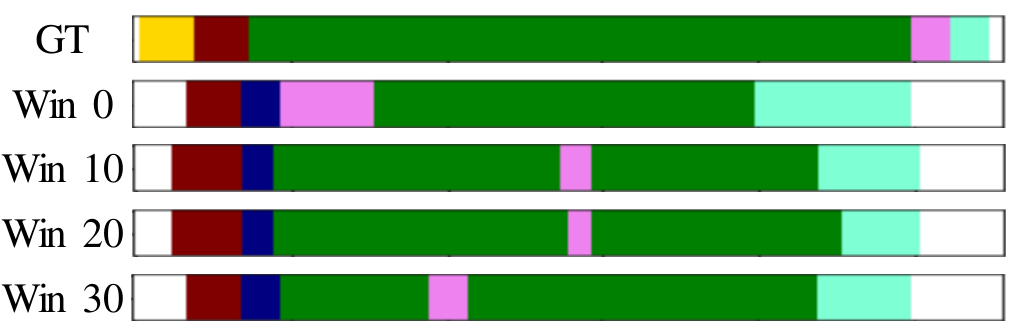}
\caption{Ground truth action sequence (\textcolor{mygold}{pour\_oil}, \textcolor{mymaroon}{crack\_egg}, \textcolor{mygreen}{fry\_egg}, \textcolor{myviolet}{take\_plate}, \textcolor{myaquamarine}{put\_egg2plate}) (top) and CDFL's action segmentations using different neighbor-window sizes on the sample test video \textit{P04\_webcam02\_P04\_friedegg} from Breakfast. The background frames are marked in white. The window size of $20$ gives the best performance.}
\label{fig:diff_win}
\end{figure}

Fig.~\ref{fig:diff_win} illustrates the CDFL's action segmentations on a sample test video from the Breakfast Action dataset using different window sizes and $L_\text{CDF}$. As can be seen, considering neighboring frames around the anchor segmentation improves performance. 

Tab.~\ref{table:Different regularization} shows how different regularization factors $\alpha$  in $L_\text{DF}$ affect our action segmentation on the Breakfast Action dataset, for different neighbor-window sizes. As expected, using small $\alpha$ in training tends to give better performance. The best accuracy is achieved with $\alpha = 0.1$ and window size of $20$.

\begin{table}[!tp]
\begin{center}
\begin{tabular}{|c|c|c|c|c|}
\hline  {\bf Breakfast} & Mof & Mof-bg & IoU & IoD\\
\hline ECTC\cite{huang2016connectionist}& 35.0 & - & - & 45.0\\
\hline HTK \cite{kuehne2016end}& 43.9 & - & 26.6 & 42.6 \\
\hline OCDC \cite{bojanowski2014weakly}& - & - & - & 23.4\\
\hline HMM/RNN \cite{richard2017weakly}& - & - & - & 47.3\\
\hline TCFPN \cite{ding2018weakly}& 53.5 & 51.7 & 35.3 & 52.3 \\
\hline D3TW \cite{chang2019d3tw} & 57.0 & - & - & 56.3\\
\hline Our \abbrmodel & {\bf 63.0} & {\bf 61.4} & {\bf 45.8} & {\bf 63.9}\\
\hline
\hline  {\bf Hollywood Ext} & Mof & Mof-bg & IoU & IoD\\
\hline ECTC\cite{huang2016connectionist}& - & - & - & 41.0\\
\hline HTK \cite{kuehne2016end}& 49.4 & - & 29.1 & 46.9 \\
\hline OCDC \cite{bojanowski2014weakly}& - & - & - & 43.9\\
\hline HMM/RNN \cite{richard2017weakly}& - & - & - & 46.3\\
\hline TCFPN \cite{ding2018weakly}& 57.4 & 36.1 & 22.3 & 39.6 \\
\hline NN-Viterbi \cite{richard2018neuralnetwork}& - & - & - & 48.7\\
\hline D3TW \cite{chang2019d3tw} & 59.4 & - & - & 50.9\\
\hline Our \abbrmodel & {\bf 64.3} & {\bf 70.8} & {\bf 40.5} & {\bf 52.9}\\
\hline
\hline {\bf 50Salads} & Mof & Mof-bg & IoU & IoD\\
\hline Our \abbrmodel & {\bf 68.0} & {\bf 65.3} & {\bf 45.5} & {\bf 58.7}\\
\hline
\end{tabular}
\end{center}
\caption{Action alignment evaluations on Breakfast, Hollywood Ext and 50Salads. The dash indicates that no results reported by prior work.}
\label{table:alignment_result}
\end{table}
\setlength{\tabcolsep}{1.4pt}

\subsection{Action Alignment}

\begin{figure}[!tp]
\centering
\includegraphics[width=1.0\linewidth]{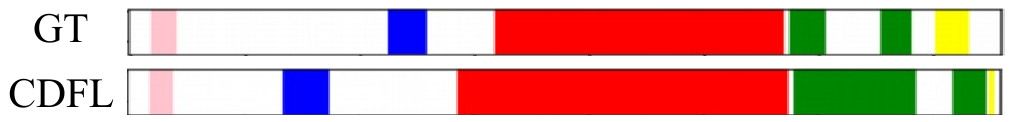}
\caption{Ground truth action sequence (\textcolor{mypink}{StandUp},\textcolor{blue}{SitDown}, \textcolor{red}{DriveCar}, \textcolor{mygreen}{OpenDoor}, \textcolor{mygreen}{OpenDoor}, \textcolor{myyellow}{HugPerson}) (top) and our action alignments (bottom) on the sample video $\textit{0261}$ from Hollywood Extend. The background frames are marked in white. \abbrmodel\ typically achieves a good action alignment.}
\label{fig:alignement}
\end{figure}

Tab.~\ref{table:alignment_result} shows that \abbrmodel \ outperforms the state-of-the-art approaches in action alignment on the three benchmark datasets. Fig.~\ref{fig:alignement} illustrates that \abbrmodel \ is good at action alignment on a sample test video from Hollywood Extended.

{\bf  Ablation Study for Action Alignment.} Tab.~\ref{table:Different Training Strategies on alignment} presents our alignment results using different loss functions as specified in Sec.~\ref{sec:CDFL}, and different neighbor-window sizes on Hollywood Ext. From the table, training with $L_{\text{DF}}$ and $L_{\text{CDF}}$ that account for invalid paths, outperforms our approach trained with $L_{\text{F}}$. In addition, taking into account neighboring frames around segmentation cuts of the initial segmentation (i.e., window size is greater than $0$) improves performance relative to the case when window size is $0$. The best performance is achieved using $L_{\text{CDF}}$ with the window sizes of $6$ in training. 

\begin{figure}
\centering
\includegraphics[width=1.0\linewidth]{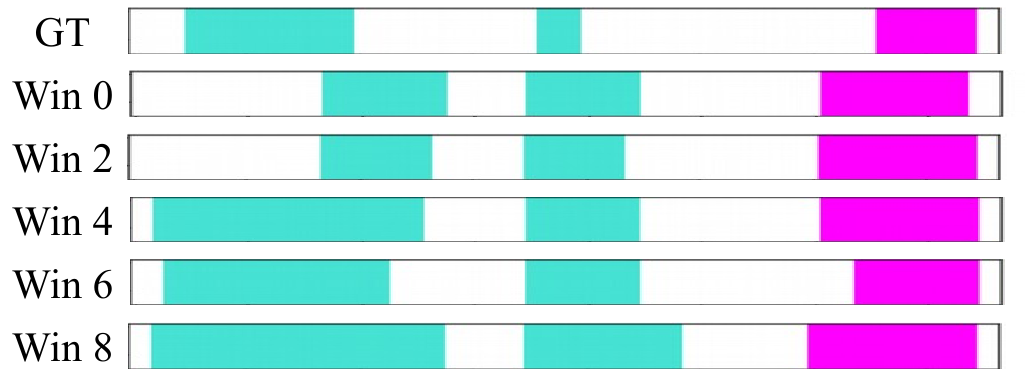}
\caption{Ground truth action sequence (\textcolor{myturquoise}{OpenDoor}, \textcolor{myturquoise}{OpenDoor}, \textcolor{mymagenta}{OpenCarDoor}) (top) and CDFL's action alignments on the sample test video \textit{0361} from Hollywood Extended, when trained using varying window sizes. The background frames are marked in white. Using \abbrmodel{} and neighbor-window size of $6$ gives the best results. }
\label{fig:diff_win_align}
\vspace{-2mm}
\end{figure}

Fig.~\ref{fig:diff_win_align} illustrates that CDFL gives good action alignment results on the sample test video from Hollywood Ext, using $L_{\text{CDF}}$ and window size of $6$ in training.

\begin{table}
\begin{center}
\begin{tabular}{|c|c|c|c|}
\hline  Window size & $L_\text{F}$ & $L_\text{DF}$& $L_\text{CDF}$\\
\hline 8 & 48.7 & 49.8 & 51.6\\
\hline 6 & 49.3 & 50.5 & ${\bf 52.9}$\\
\hline 4 & 49.0 & 50.0 & 52.0\\
\hline 2 & 48.5 & 49.5 & 50.7\\
\hline 0 & 48.7 & 49.3 & 49.8\\
\hline
\end{tabular}
\end{center}
\caption{IoD evaluations of our approach in action alignment on Hollywood Extended using different loss functions and different neighbor-window sizes in training. Using \abbrmodel{} with neighbor-window size of $6$ shows the best result.}
\label{table:Different Training Strategies on alignment}
\end{table}

\section{Conclusion}\label{sec:Conclusion}
We have extended the existing work on weakly supervised action segmentation that uses an HMM and GRU for labeling video frames by formulating a new energy-based learning on a video's segmentation graph. The graph is constructed so as to facilitate computation of loss, expressed in terms of the energy of valid and invalid paths representing candidate action segmentations. Our key contribution is the new recursive algorithm for efficiently computing the accumulated energy of exponentially many paths in the segmentation graph. Among the three loss functions that we have defined, 
and evaluated,
the CDFL -- specified to maximize the discrimination margin between valid and high-scoring invalid paths -- gives the best performance. A comparison with the state of the art on both action segmentation and action alignment tasks, for the Breakfast Action, Hollywood Extended and 50Salads datasets, supports our novelty claim that using our CDFL in training gives superior results than a loss function estimated on a single inferred segmentation, as done by prior work. Our results on both action segmentation and action alignment tasks also demonstrate advantages of considering many candidate segmentations in neighbor-windows around the initial video segmentation, and maximizing the margin between all valid and hard invalid segmentations. Our small increase in complexity relative to that of related work seems justified considering our significant performance improvements.

\noindent{\bf{Acknowledgement}}. This work was supported in part by DARPA XAI Award N66001-17-2-4029 and AFRL STTR AF18B-T002.




{\small
\bibliographystyle{ieee_fullname}
\bibliography{egbib}
}

\end{document}